\documentclass{llncs}
\usepackage{graphicx}
\usepackage{amsfonts}
\usepackage{amsmath,mathrsfs}
\usepackage{mathtools}
\usepackage{amsbsy}
\usepackage{color}
\usepackage{subcaption}
\usepackage{multirow}
\usepackage{graphicx,float,epstopdf}
\usepackage[capitalise]{cleveref}
\graphicspath{{figures/}}
\definecolor{mypink1}{rgb}{0.858, 0.188, 0.478}

\usepackage{booktabs,dcolumn,caption}
\usepackage[
  separate-uncertainty = true,
  multi-part-units = repeat
]{siunitx}

\newcolumntype{d}[1]{D{.}{.}{#1}} 

\title{Left ventricle quantification through spatio-temporal CNNs}
\author{Alejandro Debus, Enzo Ferrante}
\institute{Research institute for
signals, systems and computational intelligence, \textit{sinc(i)}, FICH-UNL/CONICET\\
(Santa Fe, Argentina)}

\begin{document}
\maketitle

\begin{abstract}
Cardiovascular diseases are among the leading causes of death globally. Cardiac left ventricle (LV) quantification is known to be one of the most important tasks for the identification and diagnosis of such pathologies. In this paper, we propose a deep learning method that incorporates 3D spatio-temporal convolutions to perform direct left ventricle quantification from cardiac MR sequences. Instead of analysing slices independently, we process stacks of temporally adjacent slices by means of 3D convolutional kernels which fuse the spatio-temporal information, incorporating the temporal dynamics of the heart to the learned model. We show that incorporating such information by means of spatio-temporal convolutions into standard LV quantification architectures improves the accuracy of the predictions when compared with single-slice models, achieving competitive results for all cardiac indices and significantly breaking the state of the art \cite{xue2018full} for cardiac phase estimation.


\end{abstract}

\section{Introduction}
In 2015, around 17.7 million people died worldwide due to heart diseases. Left ventricle (LV) quantification is a key factor for the identification and diagnosis of such pathologies \cite{karamitsos2009role}. However, the estimation of cardiac indices remains a very complex task due to its intricated temporal dynamics and the inter-subject variability of the cardiac structures. Indices such as cavity and myocardium area, regional wall thickness, cavity dimensions, among others, provide useful information to diagnose various types of cardiac pathologies. 
Cardiovascular magnetic resonance (CMR) is one of the preferred modalities for LV related studies since it is non invasive, presents high spatio-temporal resolution, has a good signal-to-noise ratio and allows to clearly identify the tissues and muscles of interest \cite{suinesiaputra2015quantification}.

\begin{figure}
\centering
\includegraphics[width=0.9\linewidth]{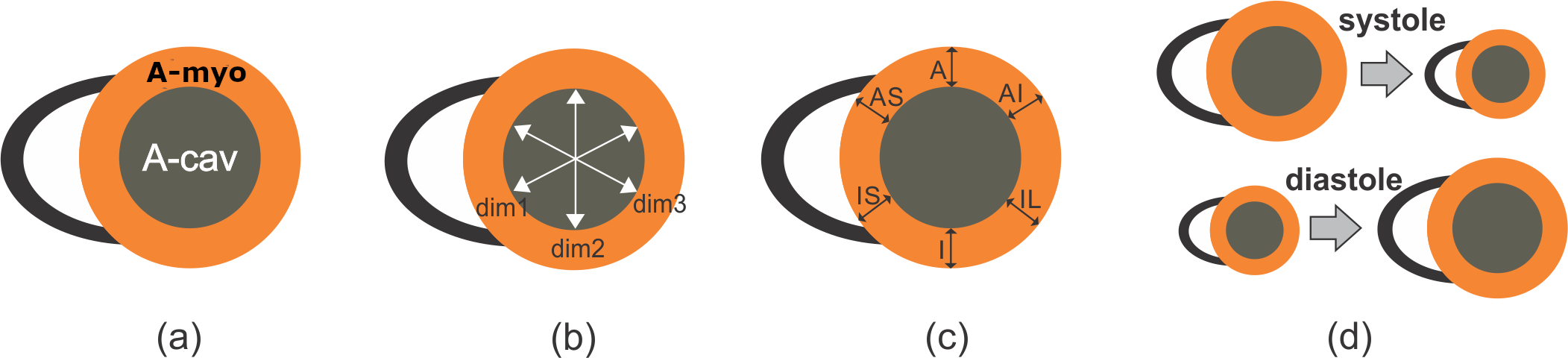}
\caption{\label{fig1}Illustration of indices of the left cardiac ventricle (based on Fig. 1 from \cite{xue2018full}). (a) Cavity area (brown) and myocardial area (orange). (b) Directional dimensions of cavity (white arrows). (c) Regional wall thicknesses. A: anterior; AS: anterospetal; IS: inferoseptal; I: inferior; IL: inferolateral; AL: anterolateral. (d) Cardiac phase (systole or diastole)}
\end{figure}

The classical approach to LV quantification consists in estimating such indices by means of automatic segmentation \cite{peng2016review,petitjean2011review,poudel2016recurrent,suinesiaputra2015quantification,tan2017convolutional,tran2016fully}. Segmentation is usually performed following supervised learning approaches, which require expert manual annotations contouring the edges of the myocardium for training. Once the segmentation is performed, the indices are computed from the resulting mask. Therefore, the accuracy of the predicted indices is conditioned on the quality of the segmentation. In this work, we follow an alternative strategy that directly estimates the indices of interest from the input image sequence. Inspired by the work of \cite{xue2017full,xue2018full,xue2017direct2}, our model is based on a convolutional neural network directly operating on images and regressing the target indices. Different from previous approaches like \cite{xue2018full} where the temporal dynamics of cardiac sequences is incorporated using recurrent neural networks (RNNs), we propose a simple but effective strategy based on the use of spatio-temporal convolutions \cite{tran2015learning}. In the context of video analysis, spatio-temporal convolutions are standard 3D convolutions that operate on spatio-temporal video volumes \cite{tan2017convolutional}. Here we employ them to process subsets of temporally contiguous CMR slices, leveraging temporal information towards improving prediction accuracy.

We investigate the use of spatio-temporal convolutions for estimating cardiac phase, directional dimensions of the cavity, regional wall thicknesses and area of cavity and myocardium under the hypothesis that such indices may be better explained when taking into account the temporal dynamics of the heart. 
We benchmark the proposed architecture using the LVQuan Challenge 2018\footnote{LVQuan Challenge website: https://lvquan18.github.io/} dataset, which provides CMR sequences with annotations for the aforementioned indices, and provide empirical evidence that incorporating the temporal dynamics of the heart through 3D spatio-temporal convolutions improves prediction accuracy when compared with single-slice models.

\section{Materials and methods}
\subsection{Architecture}
\begin{figure}[t!]
\centering
\includegraphics[width=0.9\linewidth]{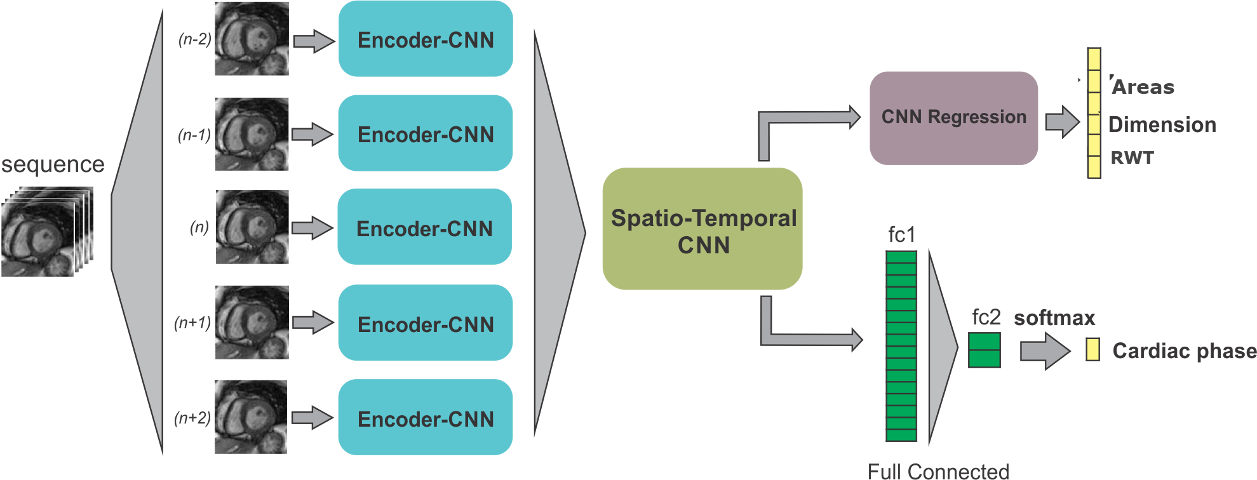}
\caption{\label{fig2}Overview of proposed architecture.}
\end{figure}
An overview of the proposed CNN architecture is presented in Figure \ref{fig2}. The network takes sequences of $\kappa$ slices and outputs the corresponding indices $only$ for the central slice. In such way, we incorporate information from the surrounding slices, easing the prediction task. 
In what follows, we describe in detail the main components of the proposed architecture. \\

\begin{figure}[t!]
\begin{subfigure}{1\textwidth}
\centering
\caption{}
\includegraphics[width=1\linewidth]{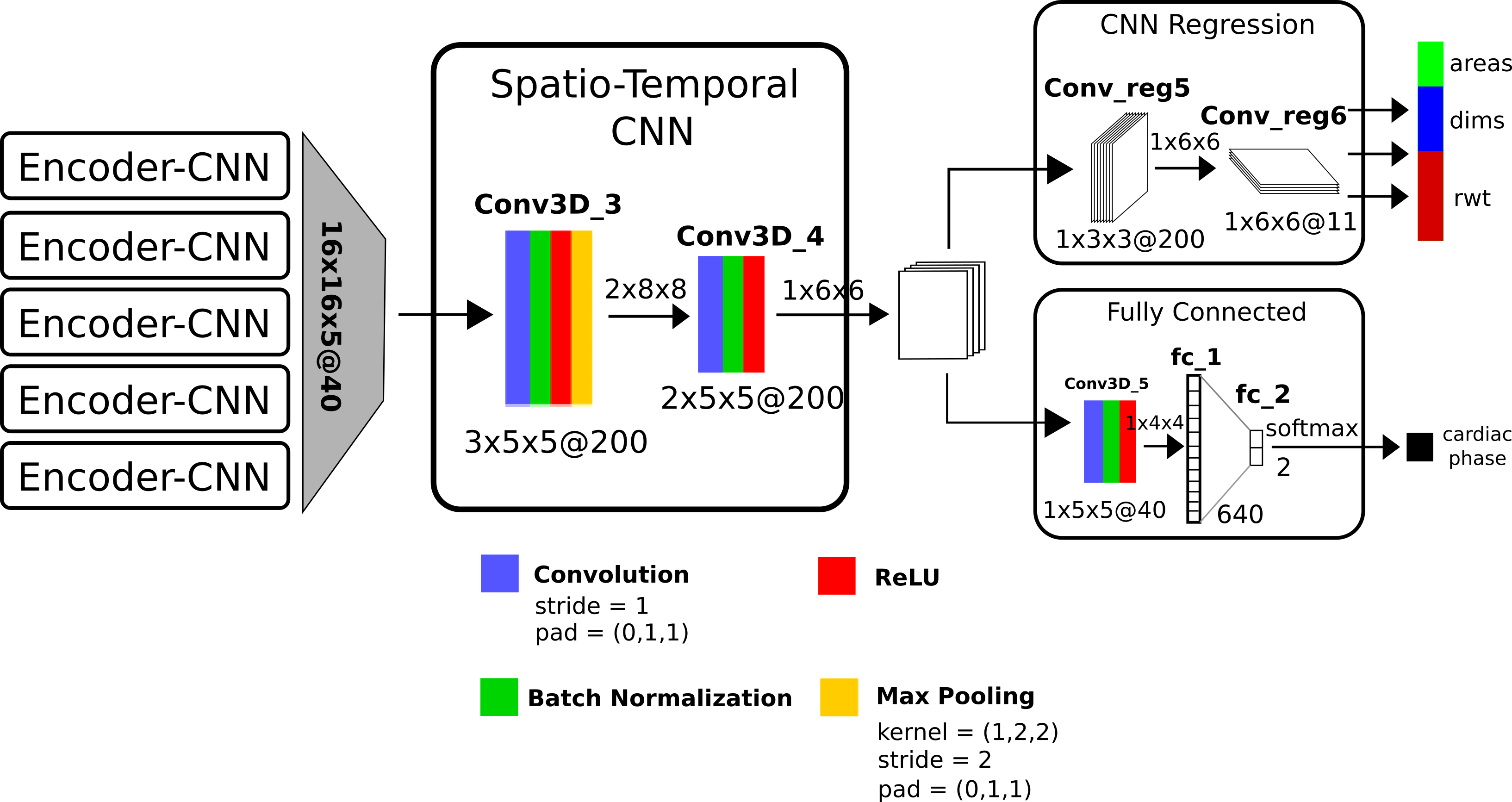}
\end{subfigure}
\begin{subfigure}{1\textwidth}
\centering
\caption{}
\includegraphics[width=0.5\linewidth]{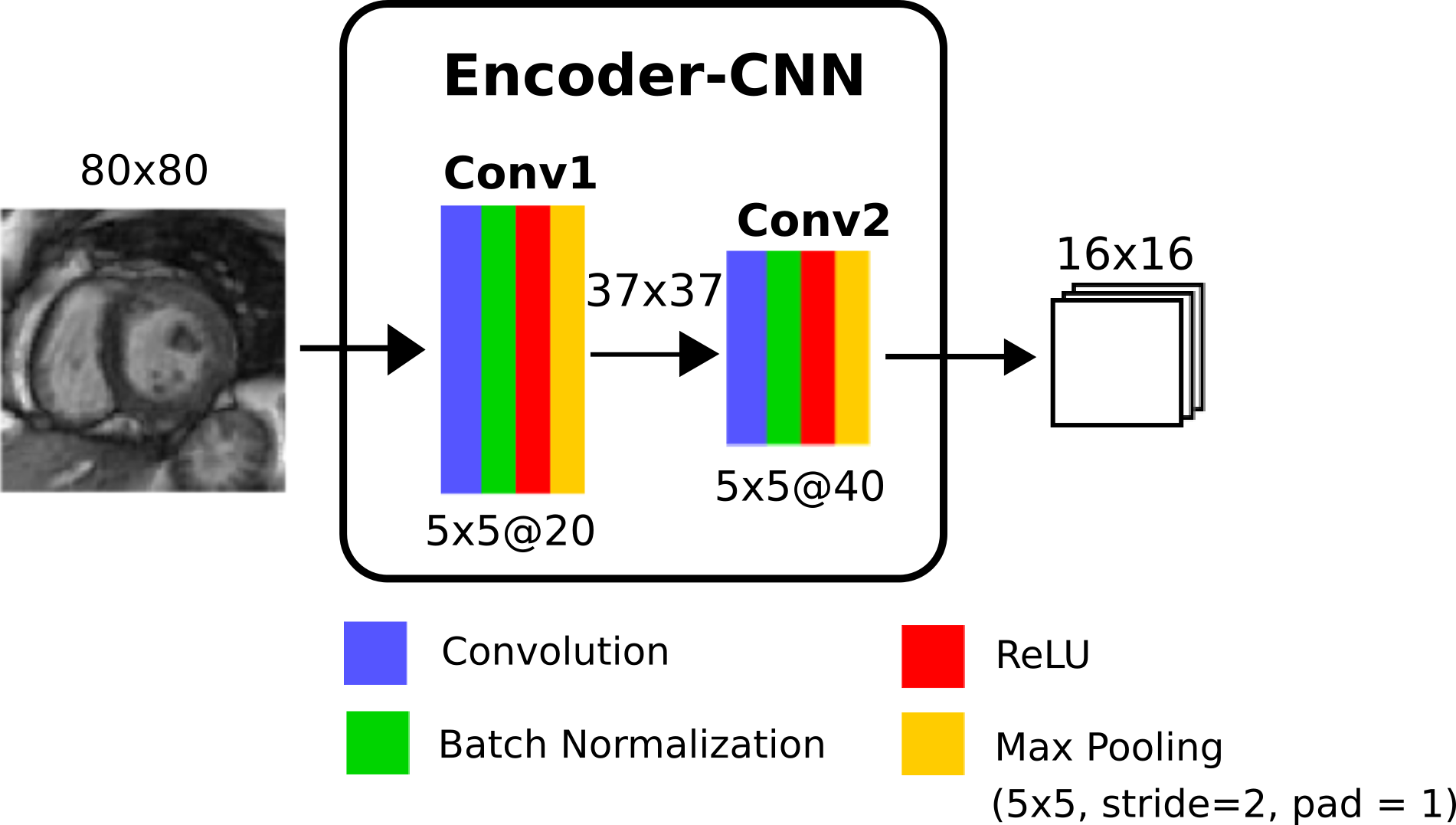}
\end{subfigure}
\caption{(a) \label{fig4}Deatiled overview of the spatio-temporal CNN based on 3D convolutions. (b) \label{fig3} Zoomed version of the  individual encoder-CNNs: for a single input slice of size 80x80 it outputs 40 filters of size 16x16 which are then fed to the spatio-temporal CNN.}
\end{figure}

\noindent \textbf{Encoder-CNN.} We use a first CNN (referred as encoder-CNN in Figures \ref{fig2} and \ref{fig3}) to extract informative features from individual slices. Inspired by \cite{xue2017full}, we designed the per-slice encoding phase using a two-layers CNN where the convolutional and pooling kernels are of size 5x5, instead of the frequently used 3x3, to introduce more shift invariance (see Figure \ref{fig3} for more details). We use ReLU activation function and batch normalization to alleviate the training process.\\

\noindent \textbf{Spatio-Temporal CNN. }After the encoding phase, the 40 filters generated for every individual encoder-CNN are used to construct a spatio-temporal volume with 40 channels per temporal slice. This volume is then processed using 3D convolutions that operate on the temporal and spatial dimensions (see Figure \ref{fig4}), producing compound feature maps that incorporate information from both of them. This module is composed of two 3D convolutional layers with kernels of size 3x5x5 and 2x5x5 when considering $\kappa=5$ slices. When considering $\kappa=1,3,7$ slices, the proposed architecture is modified by using padding in the temporal dimension ($\kappa=1,3$) and adding an extra convolution ($\kappa=7$) so that the shape of the output tensor matches 1x6x6, the size required by the CNN Regression and Fully Connected modules. ReLU activations and batch normalization are also used in this module.\\



\noindent \textbf{Final parallel branches. } After fusing the spatio-temporal features, two parallel branches are derived: (i) the first branch corresponds to a shallow CNN coupled after the spatio-temporal module, acting as a regressor of the directional dimensions, wall thickness and areas; (ii) in the second branch, a third convolutional layer is coupled to the spatio-temporal module, followed by a fully connected multi layer perceptron (MLP) with 640 neurons in the hidden layer and 2 output neurons encoding the probability for the cardiac phase (systole or diastole).\\\\

\noindent \textbf{Training procedure and loss function. }
We train the proposed network by minimizing a loss function over sets of $\kappa$ slices where annotations are provided only for the central slice. Given a set of $\kappa$ slices $\boldsymbol{x^i} =\{x_0, ... ,x_\kappa-1\}$, ground-truth annotations for the central slice $\boldsymbol{y^i}=\{ y_{dim}, y_{areas}, y_{rwt}, y_{phase} \}$ and corresponding predictions from the proposed neural network $\phi_{phase}$ and $\phi_{dim}, \phi_{areas}, \phi_{rwt}$ the loss function is defined as:

\begin{equation}
\begin{split} 
\mathcal{L}(\boldsymbol{x^i}, \boldsymbol{y^i}) = \mathcal{L}_{mse}(\phi_{areas}, y_{areas}) + \mathcal{L}_{mse}(\phi_{dim}, y_{dim}) + \\  \mathcal{L}_{mse}(\phi_{rwt}, y_{rwt}) + \mathcal{L}_{ce}(\phi_{phase}, y_{phase}) + \lambda \mathcal{L}_{reg},
\end{split} 
\end{equation}

\noindent where $\mathcal{L}_{mse}$ is the mean squared error between predictions and ground truth, $\mathcal{L}_{ce}$ is the cross-entropy loss, $\mathcal{L}_{reg}$ is the regularizer (L2 norm of the network weights) and $\lambda$ is a weighting factor. We minimize this loss using stochastic gradient descent with momentum, with mini-batches of size $s=20$.\\

\noindent \textbf{Circular hypothesis.} Since we require sets of temporally contiguous slices as input for our spatio-temporal architecture, given a sequence of $N$ slices, we adopt a circular hypothesis meaning that slice number $N-1$ is temporally followed by slice 0. This hypothesis was corroborated by visual inspection of the training dataset. Following this strategy, we generate sets of $\kappa$ slices for every sequence and use them as independent data samples. At prediction time, we employ the same hypothesis to generate the sets of test slices.

\subsection{Dataset and experimental setting}
Our method is experimentally validated using the training data provided by the LVQuan challenge 2018, composed of short axis cardiac MR images of 145 subjects. For each subject, it contains 20 frames corresponding to a complete cardiac cycle (giving a total of 2900 images in the dataset with pixel spacing ranging from 0.6836 mm/pixel to 2.0833 mm/pixel, with a mean of 1.5625 mm/pixel). The images have been collected from 3 different hospitals and subjects are between 16 and 97 years of age, with an average of 58.9 years. All cardiac images undergo several preprocessing steps (including historical tagging, rotation, ROI clipping, and resizing). The resulting images are roughly aligned with a dimension of 80x80. Epicardium and endocardium borders were manually annotated by radiologists, and used to extract the ground truth LV indices and cardiac phase. The values of regional wall thickness and the dimensions of the cavity are normalized by the dimension of the image, while the areas are normalized by the pixel number (6400).

In our experiments, we used cross validation with 3, 5 and 7 folds as suggested by the LVQuan organizers, resulting in partitions of size (49, 48, 48), (29, 29, 29, 29, 29) and (21, 21, 21, 21, 21, 20, 20) respectively. We used learning rate = 1e-4, momentum = 0.5 and $\lambda=0.005$ (these parameters were obtained by grid-search).

The model was implemented in Python\footnote{The source code for the proposed architecture is publicly available at \url{https://github.com/alejandrodebus/SpatioTemporalCNN\_lvquan}}, using PyTorch and trained in GPU.\\

\noindent \textbf{Evaluation criteria.} Pearson correlation coefficient (PCC) and Mean Absolute Error (MAE) were used to assess the performance of the algorithms for estimation of areas, dimensions and regional wall thicknesses. Error Rate (ER) was used to assess the performance for cardiac phase classification. 

\begin{equation}
PCC_{ind} = \frac{\sum_{i=1}^{N} (\phi_{ind}^{(i)}-\bar{\phi}_{ind})(y_{ind}^{(i)}-\bar{y}_{ind})}{\sqrt{\sum_{i=1}^{N}(\phi_{ind}^{(i)}-\bar{\phi}_{ind})^2\sum_{i=1}^{N}(y_{ind}^{(i)}-\bar{y}_{ind})^2}},
\end{equation}
\begin{equation}
MAE_{ind} = \frac{1}{N} \sum_{i=1}^{N}|\phi_{ind}^{(i)} - y_{ind}^{(i)}|,
\end{equation}

where $ind \in (A_1, A_2, D_1...D_3, RWT_1...RWT_6)$ , $y_{ind}$ is the ground-truth value and $\phi_{ind}$ is the estimated value.$\bar{y}_{ind}$ and $\bar{\phi}_{ind}$ are their mean values, respectively. 

\begin{equation}
ER_{phase} = \frac{\sum_{i=1}^{N} \textbf{1}( \phi_{phase}^{(i)} \not = y_{phase}^{(i)})}{N} 100\% 
\end{equation}
where $\textbf{1}()$ is the indication function, $\phi_{phase}$ and $y_{phase}$ are the estimated and ground truth value of the cardiac phase, respectively.

\section{Results and discussion}

The effectiveness of the proposed method was validated under the experimental setting discussed in Section 2.2. We measured the influence of the parameter $\kappa$ (number of contiguous slices fed to the network) for $\kappa = $ 1 (single slice), 3, 5 and 7 for the proposed spatio-temporal model based on 3D convolutions, and compare with the state of the art method recently proposed in \cite{xue2018full}. Results are presented in Table \ref{tab1} for a 5-fold cross validation setting (the same experimental setting and dataset was used in \cite{xue2018full}). Note that using sets of $\kappa=5$ slices significantly outperforms the configurations $\kappa=1,3$ for all the indices, highlighting the importance of the temporal dynamics.  However, considering $\kappa=5$ and $\kappa=7$ slices achieves a similar performance. Therefore, we consider $\kappa=5$ as enough temporal context for the remaining experiments.

In quantitative terms, we reduce the error rate from 28.45.06\% to 3.85\% for cardiac phase estimation and the MAE from 270 to 190$mm^2$, 3.18 to 2.29$mm$ and 2.62 to 1.42$mm$ in average for the areas, directional dimensions of the cavity and regional wall thickness when comparing the performance for $\kappa=1$ and $\kappa=5$ slices respectively. Moreover, considering the baseline \cite{xue2018full} we observe similar results for most indices, except for the phase, where our model improves over the state of the art by a significant margin (reducing the error rate from 8.2\% to 3.2\%)


Finally, table \ref{tab2} presents these results for 3 different cross-validation configurations (3, 5 and 7 folds) as required by the LVQuan challenge organizers, together with the results for phase, directional dimensions, regional wall thicknesses and area of cavity and myocardium obtained with the best performing spatio-temporal model ($\kappa=5$). Note that performance is consistent across folds.

\begin{table}[H]
\resizebox{1.\linewidth}{!}{
\begin{tabular}{lccccc}
\hline
                          & \multicolumn{5}{c}{Model}                                                                                                           \\ \hline
                          & $\kappa$=1      & $\kappa$=3      & $\mathbf{\kappa=5}$      & $\kappa$=7               & DMTRL \cite{xue2018full} \\ \hline
                          & \multicolumn{5}{c}{Areas $(mm^2)$}                                                                                                  \\ \hline
\multirow{2}{*}{a-cav}    & $239\pm198$     & $194\pm188$     & $181\pm130$              & $180\pm145$              & $\mathbf{172\pm148}$                      \\
                          & $0.861\pm0.053$ & $0.922\pm0.035$ & $0.940\pm0.014$          & $0.923\pm0.016$          & $\mathbf{0.943}$                          \\
\multirow{2}{*}{a-myo}    & $301\pm243$     & $223\pm179$     & $199\pm138$              & $207\pm141$              & $\mathbf{189\pm159}$                      \\
                          & $0.852\pm0.047$ & $0.892\pm0.029$ & $0.923\pm0.016$          & $0.931\pm0.017$          & $\mathbf{0.947}$                          \\
\multirow{2}{*}{average}  & $270\pm154$     & $208\pm141$     & $190\pm122$              & $193\pm115$              & $\mathbf{180\pm118}$                      \\
                          & $0.857\pm0.049$ & $0.907\pm0.033$ & $0.932\pm0.015$          & $0.927\pm0.018$          & $\mathbf{0.945}$                          \\ \hline
                          & \multicolumn{5}{c}{Dimensions ($mm$)}                                                                                               \\ \hline
\multirow{2}{*}{dim1}     & $3.05\pm2.84$   & $2.63\pm2.01$   & $\mathbf{2.27\pm1.79}$   & $2.31\pm1.81$            & $2.47\pm1.95$                             \\
                          & $0.861\pm0.031$ & $0.925\pm0.018$ & $\mathbf{0.961\pm0.012}$ & $0.952\pm0.015$          & $0.957$                                   \\
\multirow{2}{*}{dim2}     & $3.23\pm3.02$   & $2.80\pm1.89$   & $\mathbf{2.38\pm1.90}$   & $2.41\pm2.03$            & $2.59\pm2.07$                             \\
                          & $0.879\pm0.061$ & $0.932\pm0.023$ & $0.957\pm0.012$          & $\mathbf{0.961\pm0.013}$ & $0.894$                                   \\
\multirow{2}{*}{dim3}     & $3.27\pm3.12$   & $2.56\pm1.75$   & $\mathbf{2.22\pm1.78}$   & $2.23\pm1.67$            & $2.48\pm2.34$                             \\
                          & $0.912\pm0.047$ & $0.939\pm0.021$ & $\mathbf{0.963\pm0.011}$ & $0.959\pm0.010$          & $0.943$                                   \\
\multirow{2}{*}{average}  & $3.18\pm2.54$   & $2.66\pm1.75$   & $\mathbf{2.29\pm1.59}$   & $2.31\pm1.62$            & $2.51\pm1.58$                             \\
                          & $0.884\pm0.044$ & $0.932\pm0.022$ & $\mathbf{0.960\pm0.011}$ & $0.957\pm0.012$          & $0.925$                                   \\ \hline
                          & \multicolumn{5}{c}{Regional wall Thickness $(mm)$}                                                                                  \\ \hline
\multirow{2}{*}{wt1 (IS)} & $2.02\pm1.32$   & $1.89\pm1.04$   & $\mathbf{1.23\pm1.14}$   & $1.24\pm1.17$            & $1.26\pm1.04$              \\
                          & $0.625\pm0.063$ & $0.793\pm0.056$ & $0.854\pm0.014$          & $0.846\pm0.011$          & $\mathbf{0.856}$                          \\
\multirow{2}{*}{wt2 (I)}  & $2.67\pm1.69$   & $2.45\pm1.48$   & $1.44\pm1.22$            & $1.43\pm1.87$            & $\mathbf{1.40\pm1.10}$                    \\
                          & $0.618\pm0.055$ & $0.751\pm0.037$ & $0.797\pm0.011$          & $\mathbf{0.801\pm0.014}$ & $0.747$                                   \\
\multirow{2}{*}{wt3 (IL)} & $2.95\pm2.01$   & $1.74\pm1.56$   & $\mathbf{1.57\pm1.41}$   & $1.60\pm1.59$            & $1.59\pm1.29$                             \\
                          & $0.595\pm0.049$ & $0.735\pm0.033$ & $\mathbf{0.765\pm0.013}$ & $0.740\pm0.010$          & $0.693$                                   \\
\multirow{2}{*}{wt4 (AL)} & $2.77\pm1.65$   & $1.66\pm1.17$   & $1.48\pm1.13$            & $\mathbf{1.46\pm1.45}$   & $1.57\pm1.34$                             \\
                          & $0.603\pm0.052$ & $0.763\pm0.024$ & $\mathbf{0.785\pm0.022}$ & $0.782\pm0.018$          & $0.659$                                   \\
\multirow{2}{*}{wt5 (A)}  & $3.06\pm2.12$   & $1.49\pm1.35$   & $1.35\pm1.19$            & $1.39\pm1.21$            & $\mathbf{1.32\pm1.10}$                    \\
                          & $0.642\pm0.061$ & $0.808\pm0.029$ & $0.842\pm0.019$          & $\mathbf{0.851\pm0.015}$ & $0.777$                                   \\
\multirow{2}{*}{wt6 (AS)} & $2.25\pm1.72$   & $1.65\pm1.11$   & $1.46\pm1.32$            & $1.49\pm1.37$            & $\mathbf{1.25\pm1.01}$                    \\
                          & $0.651\pm0.047$ & $0.825\pm0.032$ & $0.870\pm0.015$          & $0.866\pm0.013$          & $\mathbf{0.877}$                          \\
\multirow{2}{*}{average}  & $2.62\pm2.10$   & $1.81\pm1.05$   & $1.42\pm0.65$            & $1.43\pm0.71$            & $\mathbf{1.39\pm0.68}$                    \\
                          & $0.622\pm0.054$ & $0.779\pm0.033$ & $\mathbf{0.819\pm0.015}$ & $0.814\pm0.014$          & $0.768$                                   \\ \hline
                          & \multicolumn{5}{c}{Phase (ER\%)}                                                                                                    \\ \hline
phase                     & $28.45\pm5.50$  & $14.67\pm3.65$  & $\mathbf{3.85\pm2.82}$   & $3.91\pm2.76$            & $8.2$                                     \\ \hline
\end{tabular}
}
\vspace{5mm}
\caption{Sensitivity analysis for the parameter $\kappa$ (number of neighbouring slices) when using the spatio-temporal model based on 3D convolutions with 5-folds cross validation, compared with the state of the art DMTRL proposed in \cite{xue2018full}. Note that incorporating the temporal dynamics by considering multiple slices ($\kappa=3,5,7$) makes a significant different with respect the single slice case ($\kappa=1$). However, considering $\kappa=5$ and $\kappa=7$ slices present a similar performance. Therefore, we consider $\kappa=5$ as enough temporal context for the remaining experiments. When comparing with \cite{xue2018full} we observe similar results for most indices, expept for the phase, where the proposed model breaks the state of the art significantly (from 8.2\% to 3.2\%).}
\label{tab1}
\end{table}

\newpage
\section{Conclusions}
In this work, we proposed a new CNN architecture for LV quantification that incorporates the dynamics of the heart by means of spatio-temporal convolutions. Differently from other methods that rely on more complex mechanisms (like recurrent neural networks \cite{xue2018full}) we employ simple 3D convolutions to fuse information coming from temporally contiguous CMR slices. We generated training samples following a circular hypothesis, meaning that first and last slices of the sequences are considered as temporally contiguous. Validation was performed using CRM sequences provided by the LVQuan challenge organizers. Results show that incorporating temporal information through spatio-temporal convolutions significantly boosts prediction performance for all the indices. Moreover, when compared with the RNN based model presented in \cite{xue2018full}, we observe a significant reduction in error rate for phase estimation (from 8.2\% to 3.85\%) while keeping equivalent results for the other indices. More importantly, our method achieves state of the art results employing simple 3D convolutions instead of the more complex parallel RNN and Bayesian based multitask relationship learning module proposed in \cite{xue2018full}.

In this work we incorporated the spatio-temporal dynamics by means of 3D convolutions. However, if we consider the slices as multiple channels of a standard 2D architecture, conventional 2D convolutions could also be used, reducing the complexity of the model. Moreover, temporal information encoded by inter-slice deformation fields (obtained trough deep learning based image registration methods \cite{Ferrante2018}) could also be considered to improve model performance. In the future, we plan to explore the performance of these models when compared with the proposed architecture.

\begin{table}[t!]
\setlength\tabcolsep{8pt}
\centering
\begin{tabular}{lcccccc} 
\hline
         \multicolumn{7}{c}{\textbf{N-fold cross validation as required by LVQuan Challenge}}                                  \\ 
\hline
        & \multicolumn{3}{c}{MAE}       & \multicolumn{3}{c}{PCC}                      \\ 
\hline
        & N=3 & N=5               & N=7 & N=3               & N=5 & N=7                \\ 
\hline
& \multicolumn{6}{c}{Areas $(mm^2)$}                                                    \\ 
\hline
a-cav   &  $185\pm125$   &  $181\pm130$    &  $183\pm115$   &  0.932   &  0.940   &    0.939       \\ 
a-myo   &  $204\pm143$   &   $199\pm138$   &   $198\pm145$  &  0.915   &  0.923   &    0.930       \\ 
average &   $194\pm131$  &   $190\pm122$   &  $190\pm110$   &  0.924  &   0.932   &    0.935        \\ 
\hline
        
        & \multicolumn{6}{c}{Dimensions $(mm)$}                                               \\ 
\hline
dim1    &  $2.71\pm2.11$   &   $2.27\pm1.79$   &  $2.26\pm1.82$   &  0.938   &  0.961   &   0.959         \\ 
dim2    &  $2.65\pm2.09$   &   $2.38\pm1.90$   &  $2.32\pm2.01$   &  0.926   &  0.957   &   0.954         \\ 
dim3    &  $2.51\pm2.20$   &   $2.22\pm1.78$   &  $2.24\pm1.91$   &  0.933   &  0.963   &   0.958                 \\ 
average &  $2.62\pm1.87$   &   $2.29\pm1.59$   &  $2.27\pm1.52$   &  0.932  &  0.960   &     0.957              \\ 
\hline
& \multicolumn{6}{c}{Regional wall Thickness $(mm)$}                                  \\ 
\hline
wt1 (IS)     &  $1.31\pm1.16$   &  $1.23\pm1.14$    &  $1.25\pm1.15$   &   0.831  &  0.854   &   0.857          \\ 
wt2 (I)     &   $1.58\pm1.10$  &  $1.44\pm1.22$    &  $1.43\pm1.41$   &  0.768  &  0.797   &   0.802           \\ 
wt3 (IL)     &  $1.62\pm1.22$   &  $1.57\pm1.41$    &  $1.56\pm1.56$   &   0.743  &  0.765   &   0.755           \\ 
wt4 (AL)     &  $1.60\pm1.08$   &  $1.48\pm1.13$   &  $1.50\pm1.11$   &   0.776  &  0.785  &    0.797           \\ 
wt5 (A)     &  $1.43\pm1.12$   &   $1.35\pm1.19$  &   $1.33\pm1.24$  &   0.829  &  0.842  &   0.861               \\ 
wt6 (AS)     &  $1.52\pm1.29$   &  $1.46\pm1.32$  &  $1.46\pm1.09$   &   0.857  &  0.870  &    0.873               \\ 
average &   $1.51\pm0.98$  &  $1.42\pm0.65$    &  $1.42\pm0.61$   &   0.801  &  0.819  &    0.824       \\ 
\hline
        & \multicolumn{6}{c}{Phase (ER \%)}                                                    \\ 
\hline
        & \multicolumn{2}{c}{N=3} & \multicolumn{2}{c}{N=5} & \multicolumn{2}{c}{N=7}  \\ 
\hline
phase   & \multicolumn{2}{c}{$5.10\pm3.72$}    & \multicolumn{2}{c}{$3.85\pm2.82$}    & \multicolumn{2}{c}{$3.81\pm3.05$}     \\
\hline
\vspace{5px}
\end{tabular}
\caption{Results obtained for the LVQuan challenge dataset using the proposed spatio-temporal model (areas of LV cavity and myocardium $(mm^2)$, directional dimensions $(mm)$, wall thicknesses $(mm)$ and cardiac phase) for $N$-folds cross validation with $N=3, 5$ and 7.}
\label{tab2}
\end{table}

\section*{Acknowledgements}
The present work used computational resources of the Pirayu Cluster, acquired with funds from the Santa Fe Science, Technology and Innovation Agency (ASACTEI), Government of the Province of Santa Fe, through Project AC-00010-18, Resolution Nº 117/14. This equipment is part of the National System of High Performance Computing of the Ministry of Science, Technology and Productive Innovation of the Republic of Argentina. We also thank NVidia for the donation of a GPU used for this project. Enzo Ferrante is a beneficiary of an AXA Research Fund grant.

\bibliographystyle{splncs03}
\bibliography{references}

\end{document}


\maketitle

\subsection{Multi-resolution study}
Classical non-learning based DIR algorithms usually adopt a pyramidal approach where image registration is performed at different image resolutions, going from coarser to finer, to improve the final capture range. In general, the same iterative optimization problem is solved at every level of the pyramid, the only difference being the resolution of the input images. In case of CNN-based image registration, one could ask the following question: is it possible to predict deformation fields $\mathcal{D}_i$ at resolution level $i$ using a model $\mathcal{U}_j$ trained with images from resolution level $j$? In other words, can we transfer the parameters of a learned CNN model across image resolutions? Intuition may suggest that this is possible without any decrease in accuracy, since classical algorithms based on iterative optimization usually keep the same parameters across different levels. However, in practice, we could see a decrease in performance.

To address this question, we designed the following multi-resolution study. We trained three different models $\mathcal{U}_0$, $\mathcal{U}_1$ and $\mathcal{U}_2$ by setting the factor of the initial down-sampling layer to 0, 1 and 2 respectively. Then, at test time\footnote{The CNN-based models take 0.06 secs on GPU and 0.08 secs on CPU to register a pair of images, while Elastix took 8.9, 4.0 and 2.47 secs on CPU for resolution levels 0, 1 and 2 respectively. In all the experiments we used Adam optimization, with LR = 1e-4 and $\lambda$=1e-6}, we measured the model performance on unseen images coming from the same domain, but with different resolutions. Figure \ref{fig:Multires} summarizes results for this experiment, also including MAD, DSC and CMD before registration and baseline results obtained by classical deformable registration using Elastix \cite{Klein2010} (Elastix parameters were chosen by grid search using the training data and are available online in our project website). The best performance was obtained when train and test images were coming from the same resolution. Level $l=2$ (in green) seems to be the resolution where the method achieves best performance in both datasets for most indicators. Even if in most cases the models trained in a resolution different from the one used at testing still perform decently, the best performance is always achieved when train and test resolutions agree.

\begin{figure}[t!]
  \centering
  \includegraphics[width=1.0\linewidth]{imgs/MultiresFigureLightColors_BigTitle.pdf}
  \caption{Multi-resolution study. CNN-based registration models trained/tested across multiple image resolutions (train and test images coming from the same domain). MAD (mean of absolute differences), CMD (contour mean distance) and DSC (Dice coefficient) before and after registration are provided, as well as baseline results obtained with the state-of-the-art Elastix toolbox for comparison.}
  \label{fig:Multires}
\end{figure}

\begin{figure}[t!]
  \centering
  \includegraphics[width=1.0\linewidth]{imgs/MultiresFigureLightColors_BigTitle.pdf}
  \caption{Multi-resolution study. CNN-based registration models trained/tested across multiple image resolutions (train and test images coming from the same domain). MAD (mean of absolute differences), CMD (contour mean distance) and DSC (Dice coefficient) before and after registration are provided, as well as baseline results obtained with the state-of-the-art Elastix toolbox for comparison.}
  \label{fig:Multires}
\end{figure}